%% file: main.tex
\documentclass[conference]{configs/IEEEtran}
\IEEEoverridecommandlockouts
\usepackage{cite}
\usepackage{amsmath,amssymb,amsfonts}
\usepackage{algorithmic}
\usepackage[hang]{footmisc}
\usepackage{footmisc}
\usepackage{graphicx}
\usepackage{textcomp}
\usepackage{xcolor}
\usepackage{caption}
\usepackage{subcaption}
\usepackage{multirow}
\def\BibTeX{{\rm B\kern-.05em{\sc i\kern-.025em b}\kern-.08em
    T\kern-.1667em\lower.7ex\hbox{E}\kern-.125emX}}

\begin{document}

\title{Efficient Expression Neutrality Estimation with Application to Face Recognition Utility Prediction \\
}


\author{\IEEEauthorblockN{1\textsuperscript{st} Marcel Grimmer}
\IEEEauthorblockA{NTNU\textsuperscript{*}\thanks{\textsuperscript{*}Norwegian University of Science and Technology} \\
Gj{\o}vik, Norway \\
marceg@ntnu.no}
\and
\IEEEauthorblockN{2\textsuperscript{st} Raymond N. J. Veldhuis}
\IEEEauthorblockA{UT\textsuperscript{†}\thanks{\textsuperscript{†}University of Twente} \qquad NTNU\textsuperscript{*} \\
Twente, Netherlands \qquad Gj{\o}vik, Norway\\
r.n.j.veldhuis@utwente.nl}
\and
\IEEEauthorblockN{3\textsuperscript{nd} Christoph Busch}
\IEEEauthorblockA{h\_da\textsuperscript{‡}\thanks{\textsuperscript{‡}Hochschule Darmstadt} \qquad NTNU\textsuperscript{*} \\
Darmstadt, Germany \qquad Gj{\o}vik, Norway\\
christoph.busch@ntnu.no}}

\maketitle

\input{sections/00-abstract}
\input{sections/01-introduction}
\input{sections/02-ExperimentalSettings}

\input{sections/03-ExperimentalResults}

\input{sections/04-Conclusion}

\section*{Acknowledgment}

This research work has been funded by the German Federal Ministry of Education and Research and the Hessian Ministry of Higher Education, Research, Science and the Arts within their joint support of the National Research Center for Applied Cybersecurity ATHENE.

\bibliographystyle{ieeetr}
\bibliography{references}

\end{document}

%% file: sections/00-abstract.tex
\begin{abstract}
The recognition performance of biometric systems strongly depends on the quality of the compared biometric samples. Motivated by the goal of establishing a common understanding of face image quality and enabling system interoperability, the committee draft of ISO/IEC 29794-5 introduces expression neutrality as one of many component quality elements affecting recognition performance. In this study, we train classifiers to assess facial expression neutrality using seven datasets. We conduct extensive performance benchmarking to evaluate their classification and face recognition utility prediction abilities. Our experiments reveal significant differences in how each classifier distinguishes \textit{neutral} from \textit{non-neutral} expressions. While Random Forests and AdaBoost classifiers are most suitable for distinguishing neutral from non-neutral facial expressions with high accuracy, they underperform compared to Support Vector Machines in predicting face recognition utility.
\end{abstract}

\begin{IEEEkeywords}
Face Recognition, Quality Assessment, Classification, Machine Learning
\end{IEEEkeywords}

%% file: sections/01-introduction.tex
\section{Introduction}
\label{sec:introduction}

Nowadays, identity verification through face recognition (FR) is an essential component of various applications, ranging from smartphone unlocking to border control~\cite{EU-Regulation-EES-InternalDocument-2017} or forensics~\cite{EU-Regulation-2017-2226-on-EES-171130}\cite{EU-ImplementingDecision-2019-329-on-EES-SampleQuality-190225}. Especially in security-relevant applications, recognition accuracy must comply with high standards~\cite{FRONTEX-BorderControl-BestPractices-InternalDocument-2015}. However, as shown by previous studies~\cite{schlett2022face}, a strong dependency exists between \textit{face image quality} and \textit{recognition performance}. Therefore, it is crucial to establish a unified understanding of face image quality to enable the exchange of facial images across biometric systems while preserving the same recognition performance (\textit{system interoperability}~\cite{EU-Regulation-2019-817-on-Interoperability-Framework-190520}). To address this issue, the current committee draft of \textit{ISO/IEC 29794-5}~\cite{ISO-IEC-29794-5-CD3-FaceQuality-231018} aims to standardize face image quality.

\subsection{Face Image Quality Assessment}

In this work, we use the term \textit{biometric quality} to denote the standardized concept of \textit{utility}~\cite{ISO-IEC-29794-1-QualityFramework-2023} that reflects the predicted positive or negative contribution of an individual sample to the overall performance of a biometric system. Specifically, biometric quality is further categorized as \textit{unified quality} or \textit{component quality}. \textit{Unified quality scores} evaluate the overall face image quality to anticipate the recognition outcome, considering all factors of variation and their interrelations. In contrast, \textit{quality components} measure the impact of each individual \textit{capture-} or \textit{subject-related} \textit{quality element} on the recognition performance. In practical applications, face image quality (FIQA) algorithms are used to filter out low-quality samples (\textit{i.e.}, unified quality) and offer subjects actionable feedback (\textit{i.e.}, component quality) when capturing their facial images. Within this setup, unified and component quality algorithms complement each other to improve the security and convenience of the biometric system.        

\subsection{Quality Component: Expression Neutrality}
\input{figures/intro-img}

As demonstrated in prior studies~\cite{pena2021facial}\cite{damer2018crazyfaces}, extreme facial expressions significantly impact recognition performance. This is caused by the increased intra-identity variation, corresponding with a higher likelihood of false non-match decisions. To address this issue, the current committee draft of the \textit{International Organization for Standardization} (ISO) and the \textit{International Electrotechnical Commission} (IEC) 29794-5~\cite{ISO-IEC-29794-5-CD3-FaceQuality-231018} introduces \textit{facial expression neutrality} as a subject-related quality component element. Hence, the ISO/IEC assumption is that facial images with neutral expressions are optimal for extracting identity-relevant information, whereas deviations from expression neutrality deteriorate the utility. Moreover, measuring the deviation from expression neutrality is essential for many applications, such as passport registrations, where maintaining a neutral facial expression is compulsory. Definitions for canonical face images, as they are formulated in ICAO 9303~\cite{ICAO-9303-p10-2021}, which, in turn, refers to ISO/IEC 39794-5~\cite{ISO-IEC-39794-5-G3-FaceImage-191015}, require such neutral expression.  

Modelling a function that maps facial images to their respective expression neutrality measures is non-trivial due to their complex relationship. Further, the perception of facial expressions varies across observers, depending on individual factors such as head shapes or demographics. To address this challenge, NeutrEx~\cite{Grimmer-Neutrex-IJCB-2023} was introduced as a quality measure that quantifies neutrality deviation based on \textit{3D Morphable Face Models} (3DMM). Despite its strong predictive performance, NeutrEx requires encoding facial images into the 3DMM parameter space, making it computationally expensive and impractical for biometric systems with high throughput rates on the one hand or for mobile devices on the other hand.

\subsection{Contribution \& Paper Structure}

This work proposes an efficient alternative to NeutrEx that complies with ISO/IEC CD3 29794-5~\cite{ISO-IEC-29794-5-CD3-FaceQuality-231018}. Specifically, we extract intermediate feature layers from a pre-trained facial expression recognition model~\cite{Savchenko-HSE-SISY-2021}. Assuming these features capture information relevant to facial expressions, we train two-class classifiers to distinguish between samples with \textit{neutral} vs \textit{non-neutral} facial expressions (see Figure~\ref{fig:intro-imgs}). Subsequently, we redefine the classifier's neutrality confidence scores as expression neutrality measures compliant with ISO/IEC CD3 29794-5. On this foundation, we extensively compare the FR utility prediction performance across various feature combinations and two-class classifiers. In summary, the contributions can be summarised as follows:

\begin{itemize}
    \item We leverage the intermediate layers of two lightweight expression recognition models \cite{Savchenko-HSE-SISY-2021} ($5.3$M and $9.3$M parameters) to derive an expression neutrality measure compliant with ISO/IEC CD3 29794-5~\cite{ISO-IEC-29794-5-CD3-FaceQuality-231018}.
    \item We compare the FR utility prediction and classification performance based on three machine learning algorithms: Support Vector Machine (SVM), Random Forest, and AdaBoost.
    \item We extensively train and evaluate our proposed classifiers on facial images from 8 open source datasets (see overview in Table~\ref{tab:datasets}) with a diverse range of facial expressions.
\end{itemize}

%% file: figures/intro-img.tex
\begin{figure}
\centering
\includegraphics[width=\linewidth]{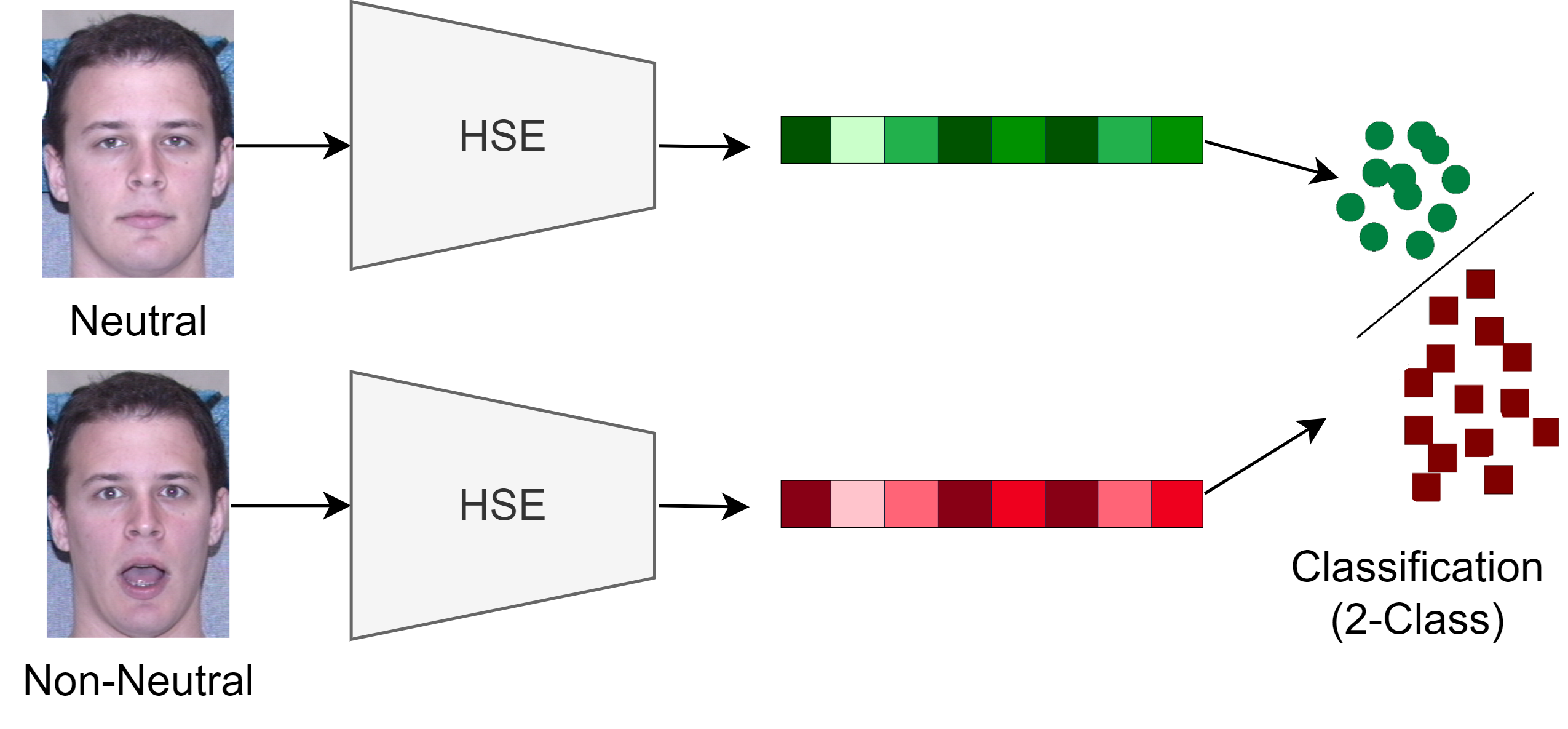}
\caption{Overview of our two-class classification approach utilizing features extracted from a pre-trained expression recognition model (HSE~\cite{Savchenko-HSE-SISY-2021}). We redefine the confidence scores of the \textit{neutral} class as an efficient measure for facial expression neutrality.}
\label{fig:intro-imgs}
\end{figure}

%% file: sections/02-ExperimentalSettings.tex
\section{Proposed Expression Neutrality Estimators}
\label{sec:experimentalSetup}

\input{figures/feature-extraction}

This section introduces the foundational expression recognition models used to extract the features from which our expression neutrality classifiers learn. Building upon these features, we describe how training two-class classifiers to distinguish between \textit{neutral} vs \textit{non-neutral} facial expressions can be applied for expression neutrality estimation. Finally, we provide an overview of all training and evaluation datasets supporting our experimental findings.   

\input{figures/feature-combination}

\subsection{Feature Extraction and Combination}

\input{tables/datasets}

One prerequisite for assessing the deviation from expression neutrality involves the modelling of patterns associated with facial expression neutrality. Defining the boundaries within which a facial expression can be classified as \textit{neutral} is a non-trivial challenge due to the inherent variation in human faces and the continuous spectrum on which facial expressions exist. To establish a data-driven definition, we extract expression-relevant features from pre-trained expression recognition models~\cite{Savchenko-HSE-SISY-2021}, assuming them to encapsulate the necessary information for classifying facial images into \textit{neutral} vs \textit{non-neutral}.

\input{figures/DET-all}

Specifically, we leverage two pre-trained models from Savchenko~\cite{Savchenko-HSE-SISY-2021}, denoted as \textit{HSE-1} and \textit{HSE-2}, building upon the EfficientNet-b0 and EfficientNet-b2 architectures~\cite{Tan-EfficientNet-PMLR-2021}. These models are trained to classify facial images into eight facial expressions associated with the following human emotions: Anger, contempt, disgust, fear, happiness, Neutrality, sadness, or surprise. Our experiments are based on features extracted from HSE-1 and HSE-2, chosen for their competitive classification performance and computational efficiency, with only $5.3$M and $9.3$M model parameters, respectively. In contrast, the alternative NeutrEx~\cite{Grimmer-Neutrex-IJCB-2023} approach involves two encoders, each with $25M$ parameters, limiting their applicability on edge devices significantly. 
 
Figure~\ref{fig:feature-extraction} illustrates the features extraction mechanism. Specifically, we extract the last feature layer embeddings $\mathcal{F}_{\text{HSE-1}}\in \mathbb{R}^{1280}, \mathcal{F}_{\text{HSE-2}}\in \mathbb{R}^{1408}$ (\textit{continuous bars}), assuming they encapsulate additional information complementary to the final Softmax output, which are represented by the \textit{discrete bars} for the eight emotion classes. By operating on $\mathcal{F}_{\text{HSE-1}}$ and $\mathcal{F}_{\text{HSE-2}}$, we anticipate the continuous nature of facial expressions that cannot be mapped to a discrete number of emotional states in an injective manner. Ultimately, based on the extracted features shown in Figure~\ref{fig:feature-extraction}, we assess the performance of our proposed expression neutrality estimators across various feature combinations shown in Figure~\ref{fig:feature-combination}.

Initially, we assess the classification accuracy and utility prediction performance of our two-class classifiers solely trained on $\mathcal{F}_{\text{HSE-1}}$ and $\mathcal{F}_{\text{HSE-2}}$ (classifiers denoted as \textit{HSE-1} and \textit{HSE-2}). Subsequently, we explore the impact of concatenating the intermediate feature embeddings with their corresponding Softmax values (classifiers denoted as \textit{HSE-1-C} and \textit{HSE-2-C}). Also, we investigate the synergy between the features of \textit{HSE-1} and \textit{HSE-2} to determine whether the two expression recognition models have learned complementary information beneficial to our applications (classifiers denoted as \textit{HSE-1-2} and \textit{HSE-1-2-C}).

\subsection{Expression Neutrality Estimation}
\label{sec:expression-neutrality-estimation}

To quantify the deviation from expression neutrality and derive a component quality measure suitable for ISO/IEC CD3 29794-5, we develop two-class classifiers to predict whether a presented facial image is classified as \textit{neutral} or \textit{non-neutral}. Specifically, we compute the expression neutrality measure by utilizing the classifier's confidence in which a facial image is associated with the \textit{neutral} expression class. By following this strategy, we hypothesize that lower neutrality confidence scores correspond with a higher deviation from expression neutrality.    

To evaluate this hypothesis and investigate how expression neutrality correlates with biometric performance, we develop three traditional classifiers\footnote{All classifiers are trained with OpenCV and can be accessed here: URL will be inserted with paper accept} on each of the six training feature combinations introduced in Figure~\ref{fig:feature-extraction}: Support Vector Machines\footnote{Hyperparameters: Kernel = "rbf", C = 3, Gamma = 0.002} (SVM), Random Forests\footnote{Hyperparameters: TermCriteria = (3, 75, 0.05), ActiveVarCount = 100, MinSampleCount = 12, MaxDepth = 25}, and AdaBoost\footnote{Hyperparameters: BoostType = 0 (discrete), WeakCount = 8000, WeightTrimRate = 0.9, MinSampleCount = 12, MaxDepth = 50} Classifiers.
 
\subsection{Training and Evaluation Datasets}
\label{sec:datasets}

The training and performance benchmarking of our classifiers are based on a diverse range of datasets to support the classifiers in generalizing across various facial expressions, head shapes, and demographic attributes. As shown in Table~\ref{tab:datasets}, we construct our training and evaluation subsets from eight distinct facial image databases, comprising the following facial expressions: Anger, Contempt, Disgust, Fear, Happiness, Neutral, Sadness, Surprise. To facilitate the classifiers in learning expression-relevant patterns without distortion from other factors of variation, we constrain each face to be fully visible (\textit{i.e.}, no extreme yaw or pitch angles).  

\input{figures/EDC-all}
\input{figures/stacked-area-plots}

For the training of the two-class classifiers, each expression other than \textit{neutral} is labelled as \textit{non-neutral}. The training dataset is balanced in terms of neutral and non-neutral facial images to facilitate equal learning of patterns from both classes by the classifiers. Additionally, we split a validation subset (30\% validation, 70\% training) for tuning the classifier-specific hyperparameters to prevent overfitting during training. Although the validation subset is randomly selected, the included identities are distinct from those in the training set. Finally, we define a single evaluation dataset composed of FEAFA+~\cite{GAN-FEAFA+-ICDIP-2022} (Subject 1-9) and MUG~\cite{Aifanti-MUG-WIAMIS-2010} to conduct the final benchmarking of the classifiers in Section~\ref{sec:experimental-results}.

%% file: figures/feature-extraction.tex
\begin{figure}
\centering
\includegraphics[width=0.7\linewidth]{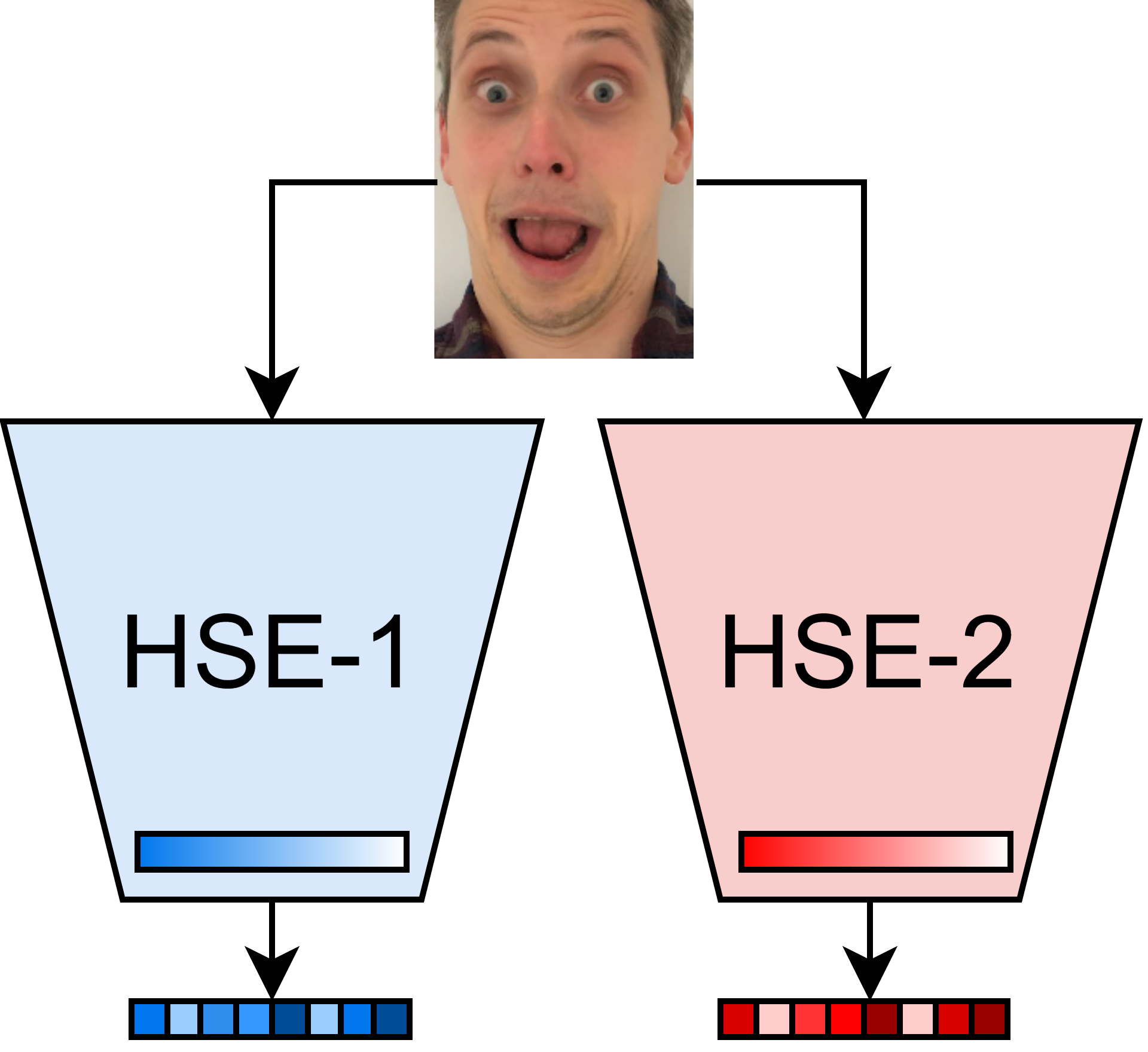}
\caption{Illustration of our feature extraction based on the expression recognition models of Savanchenko~\cite{Savchenko-HSE-SISY-2021}: HSE-1 (EfficientNet-b0~\cite{Tan-EfficientNet-PMLR-2021} and HSE-2 (EfficientNet-b2~\cite{Tan-EfficientNet-PMLR-2021}). The discrete bars represent the final Softmax scores, while the continuous bars denote the intermediate feature embeddings $\mathcal{F}_{\text{HSE-1}}\in \mathbb{R}^{1280}$, $\mathcal{F}_{\text{HSE-2}}\in \mathbb{R}^{1408}$. Figure~\ref{fig:feature-combination} demonstrates how the features are combined to train our classifiers.}
\label{fig:feature-extraction}
\end{figure}

%% file: figures/feature-combination.tex
\begin{figure}
\centering
\includegraphics[width=\linewidth]{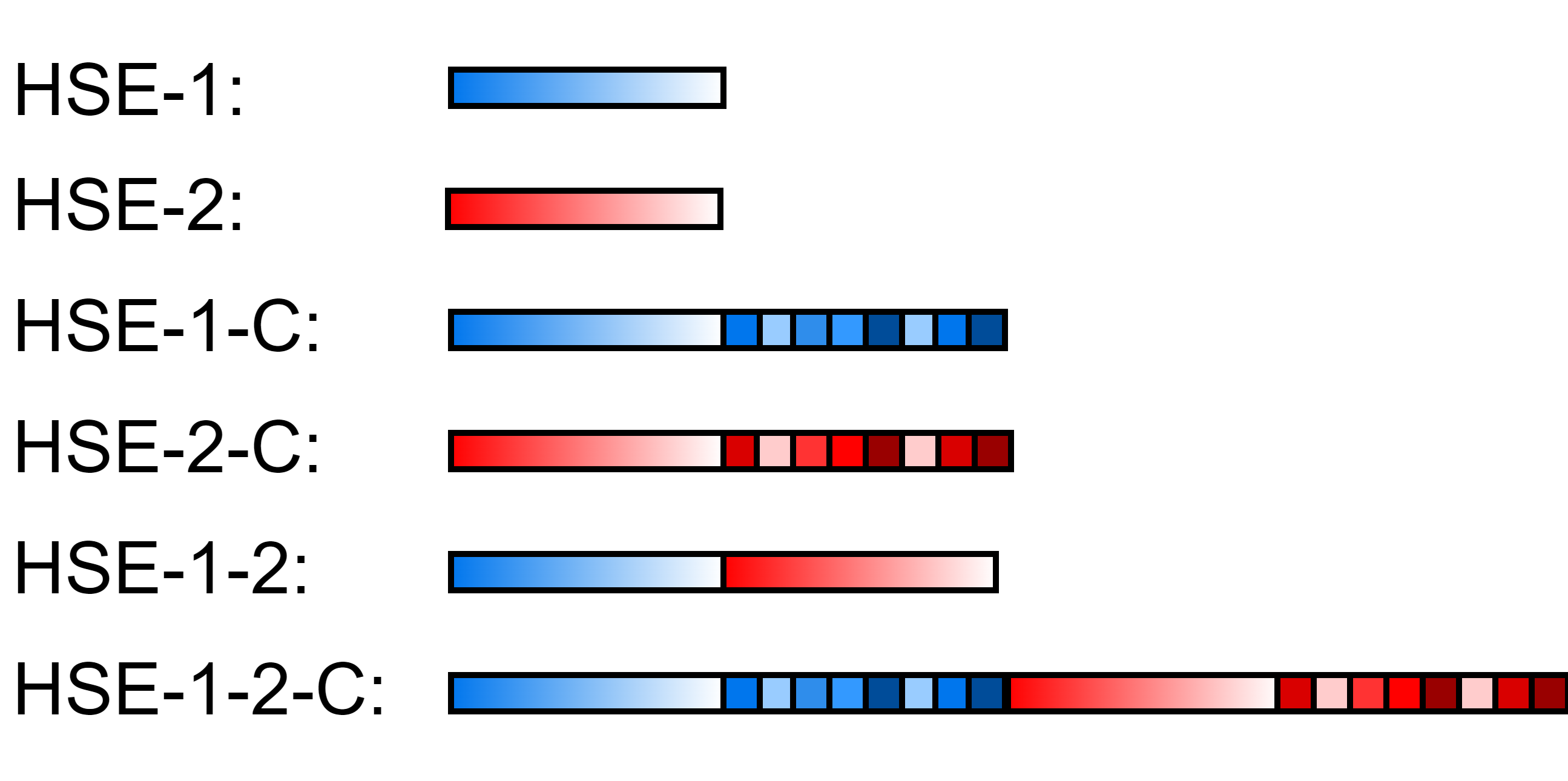}
\caption{Overview of feature combinations employed in training our two-class classifiers to distinguish \textit{neutral} from \textit{non-neutral} facial expressions. The illustrated features correspond to the feature extraction and color scheme depicted in Figure~\ref{fig:feature-extraction}.}
\label{fig:feature-combination}
\end{figure}

%% file: tables/datasets.tex
\begin{table*}[]
\caption{Overview of our training and evaluation datasets.}
\resizebox{\textwidth}{!}{%
\begin{tabular}{llccl}
\hline
                                     & \textbf{Dataset}        & \multicolumn{1}{l}{\textbf{Subjects}} & \multicolumn{1}{l}{\textbf{Images}} & \textbf{Expressions}                                                                      \\ \hline
\multirow{7}{*}{\textbf{Training}}   & CelebA-HQ~\cite{Karras-CelebAHQ-arxiv-2017}               & 238                                   & 238                                 & Contempt, Sadness                                                                         \\
                                     & CFD~\cite{Shi-CFD-TITS-2016}                     & 314                                   & 1441                                & Anger, Fear, Happiness, Neutral                                                           \\
                                     & CK+~\cite{Lucey-CK+-CVPRW-2010}                     & 123                                   & 920                                 & \multicolumn{1}{c}{Anger, Contempt, Disgust, Fear, Happiness, Neutral, Sadness, Surprise} \\
                                     & FEAFA+\cite{GAN-FEAFA+-ICDIP-2022} (Subject 10-127) & 118                                   & 2691                                & Neutral, Non-Neutral                                                                      \\
                                     & FFHQ~\cite{Karras-FFHQ-CVPR-2019}                    & 227                                   & 227                                 & Contempt, Neutral                                                                         \\
                                     & FRGCv2~\cite{Phillips-FRGC-CVPR-2005}                  & 530                                   & 967                                 & Neutral                                                                                   \\
                                     & Multi-PIE~\cite{Gross-MultiPie-IVC-2010}               & 328                                   & 3086                                & Disgust, Happiness, Neutral, Squint, Surprise                                             \\ \hline
\multirow{2}{*}{\textbf{Evaluation}} & FEAFA+~\cite{GAN-FEAFA+-ICDIP-2022} (Subject 1-9)    & 9                                     & 280                                 & Neutral, Non-Neutral                                                                      \\
                                     & MUG~\cite{Aifanti-MUG-WIAMIS-2010}                     & 52                                    & 341                                 & Anger, Disgust, Fear, Happiness, Sadness, Surprise                                        \\ \hline
\end{tabular}%
}
\label{tab:datasets}
\end{table*}

%% file: figures/DET-all.tex
\begin{figure*}
\centering
\setlength{\tabcolsep}{1pt}
\begin{tabular}{ccc}
SVM & Random Forest & AdaBoost  \\
\includegraphics[width=.24\linewidth]{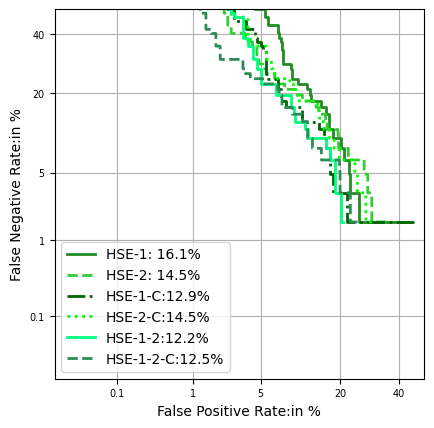} & \includegraphics[width=.24\linewidth]{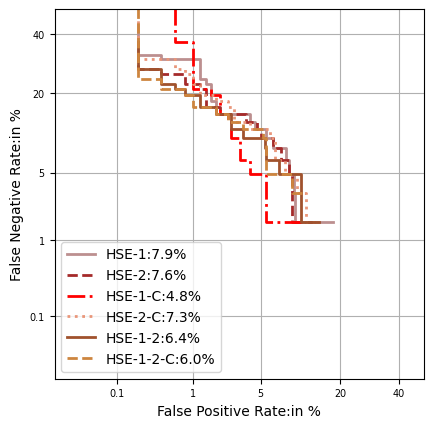} & \includegraphics[width=.25\linewidth]{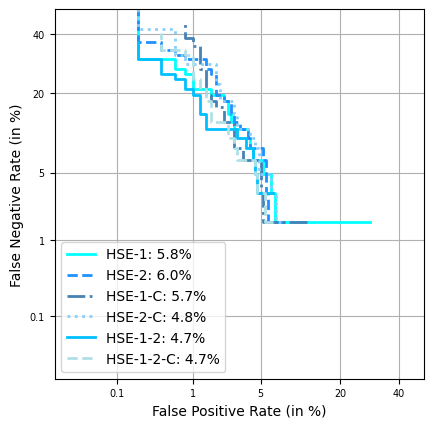} 
\end{tabular}
\caption{DET curves illustrating the classification performance benchmarking of the classifiers across six feature combinations conducted on the evaluation dataset.}
\label{fig:DET-all}
\end{figure*}

%% file: figures/EDC-all.tex
\begin{figure*}
\centering
\setlength{\tabcolsep}{1pt}
\begin{tabular}{ccc}
SVM & Random Forest & AdaBoost  \\
\includegraphics[width=.24\linewidth]{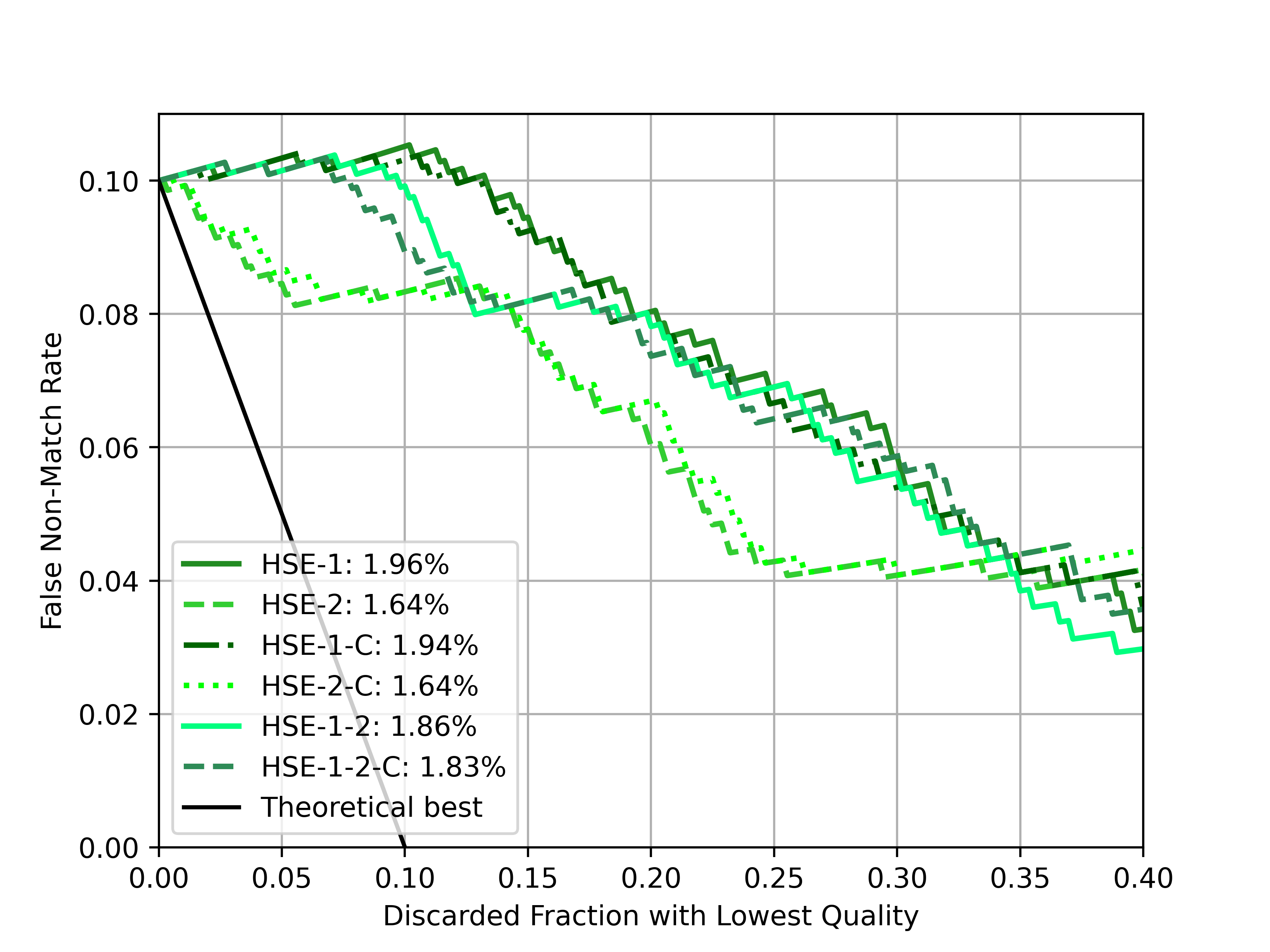} & \includegraphics[width=.24\linewidth]{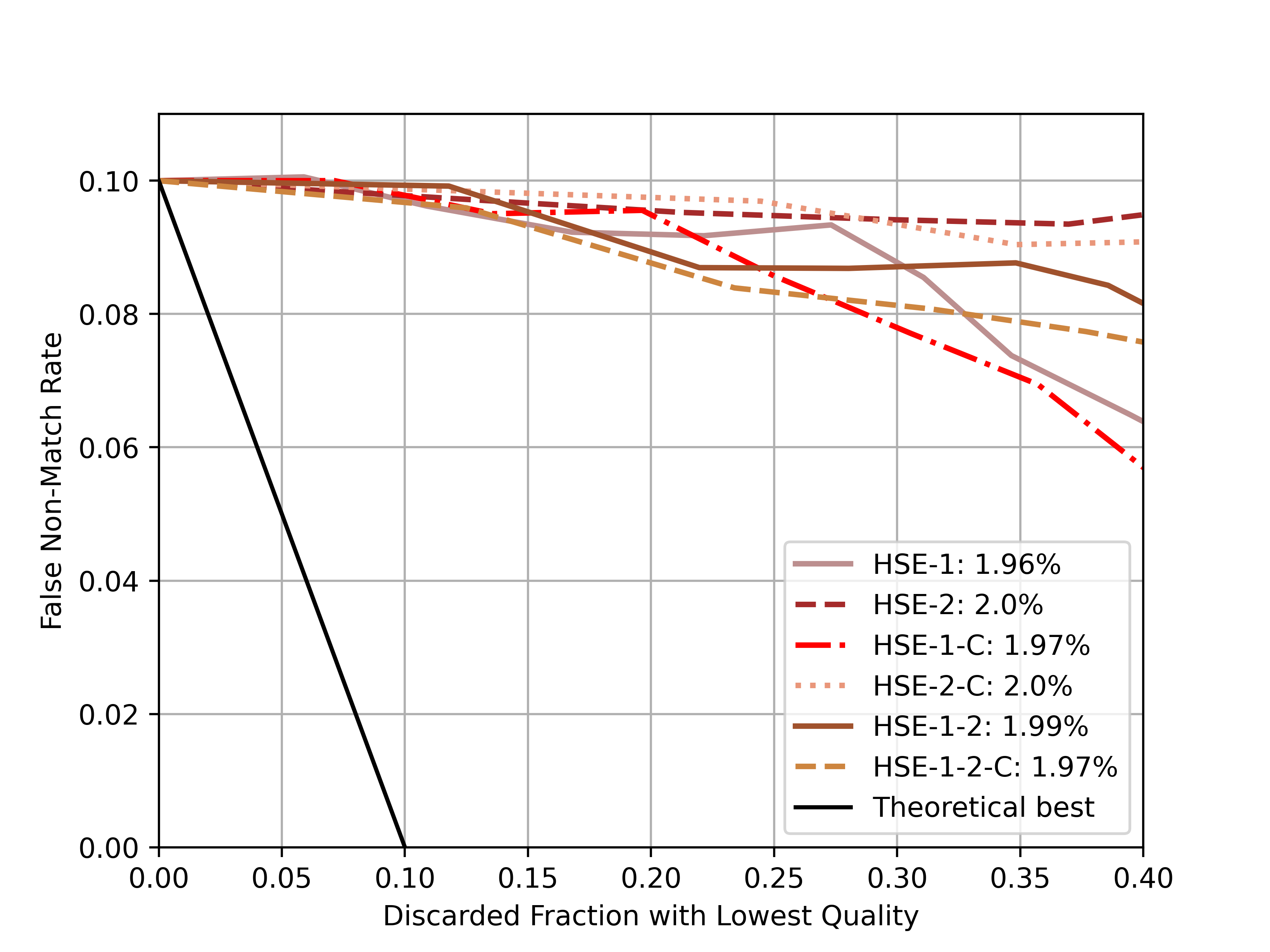} & \includegraphics[width=.25\linewidth]{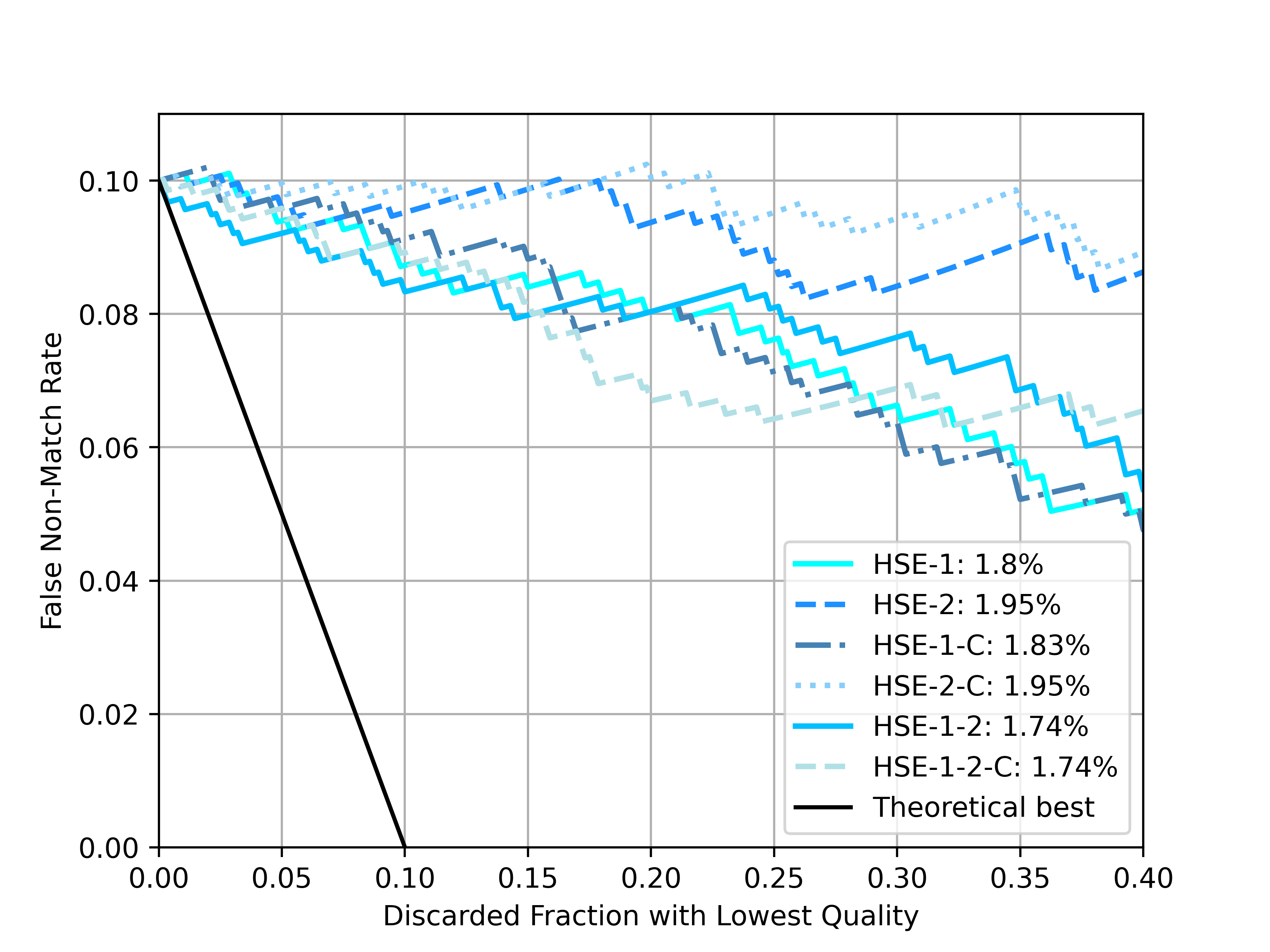}  

\end{tabular}
\caption{EDC curves depicting the FR utility prediction benchmarking of the classifiers across six feature combinations conducted on the evaluation dataset.}

\label{fig:EDC-all}
\end{figure*}

%% file: figures/stacked-area-plots.tex
\begin{figure*}
\centering
\begin{tabular}{ccc}
SVM & Random Forest & AdaBoost  \\
\includegraphics[width=.32\linewidth]{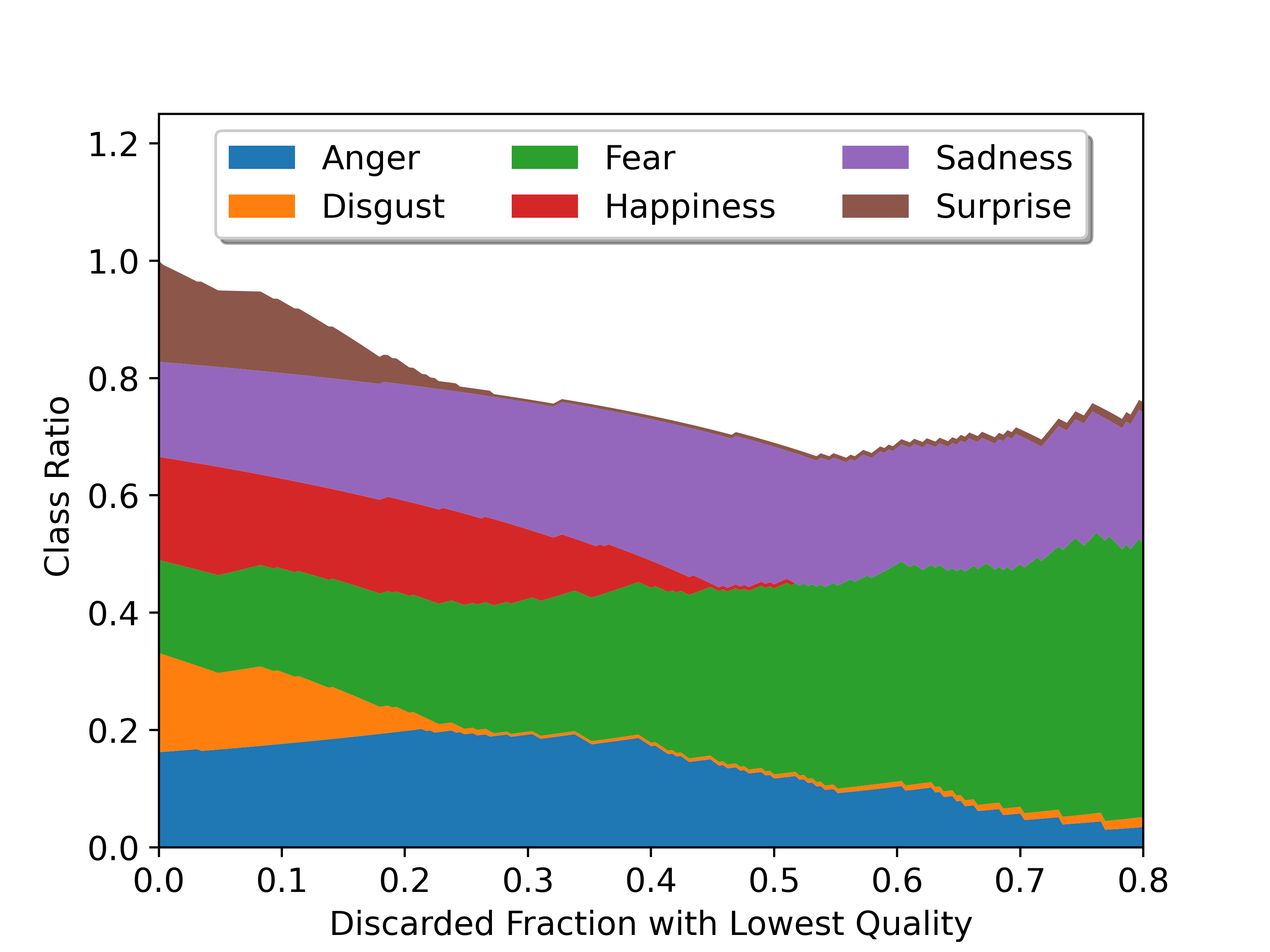} & \includegraphics[width=.32\linewidth]{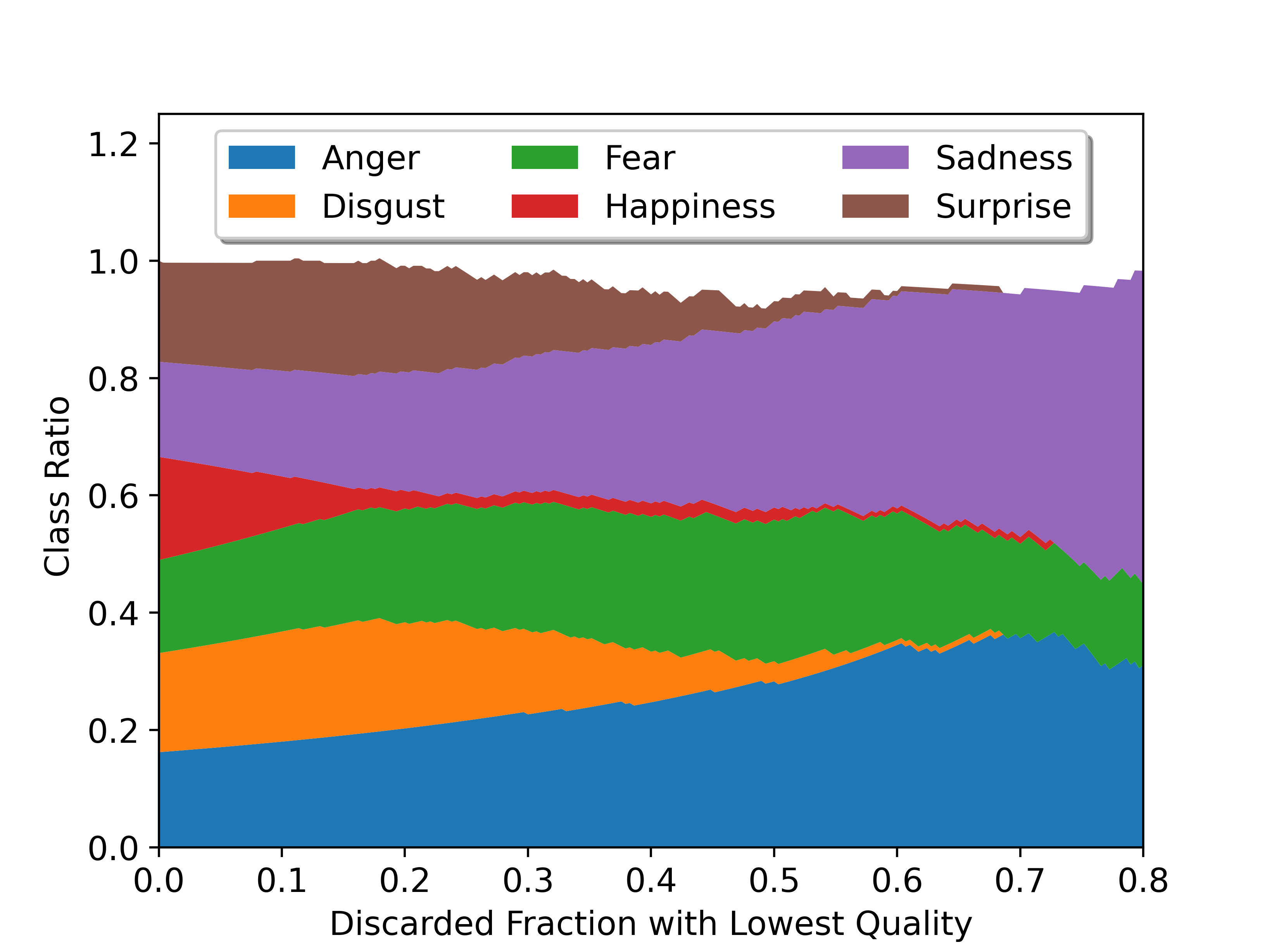} & \includegraphics[width=.32\linewidth]{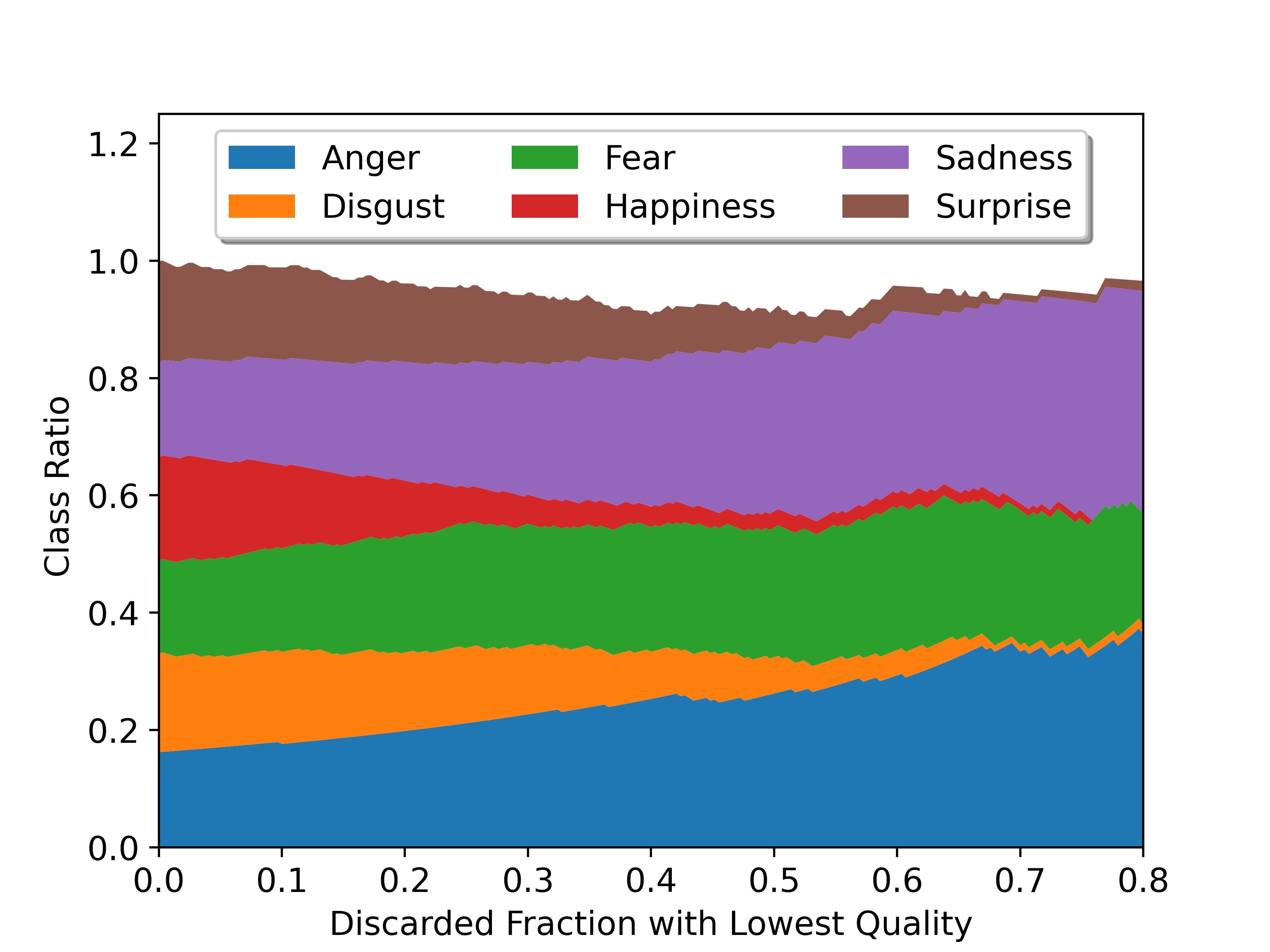} \\
\end{tabular}
\caption{Visualization of expression class proportions through successive discarding of lowest-utility facial images from the MUG~\cite{Aifanti-MUG-WIAMIS-2010} evaluation subset, estimated on their deviation from expression neutrality.}
\label{fig:stacked-area-plots}
\end{figure*}

%% file: sections/03-ExperimentalResults.tex
\section{Experimental Results}
\label{sec:experimental-results}

\subsection{Expression Classification}
\label{sec:expression-classification}

We assess the classification performance of SVM, Random Forest, and AdaBoost classifiers as introduced in Section~\ref{sec:expression-neutrality-estimation} across six feature combinations (see Figure~\ref{fig:feature-combination}). In Figure~\ref{fig:DET-all}, we compare the classification performance using \textit{Detection Error Trade-Off} (DET) curves, which visualize the trade-off between \textit{False Positive Rate} (FPR) and \textit{False Negative Rate} (FNR) on iterating thresholds. 

From the DET curves, we observe that Random Forest (\textit{HSE-1-C}) and AdaBoost (\textit{e.g.}, HSE-2-C) classifiers perform similarly, significantly surpassing the best-performing SVM (HSE-1-2). Further insights reveal that enriching features $\mathcal{F}_{HSE-1}$ and $\mathcal{F}_{HSE-2}$ with their respective Softmax scores yield a minor improvement in classification performance. Interestingly, combining features from \textit{HSE-1} and \textit{HSE-2} results in only marginal reductions in the equal error rates (EER), indicating inherent similarities in the learned patterns. Consequently, the operational efficiency can be increased by focusing on either HSE-1 or HSE-2 solely.

\subsection{Utility Prediction}
\label{sec:utility-prediction}

The second part of our study compares the performance of our expression neutrality measures in predicting FR utility. We analyze how sorting out samples with low-utility, caused by deviations from expression neutrality, affects the authentication performance of a FR system~\cite{Meng-MagFace-CVPR-2021}. To address this question, we use \textit{Error-vs-Discard Characteristic} (EDC) curves, adhering to the evaluation guidelines outlined in \textit{ISO/IEC CD3 29794-5}\cite{ISO-IEC-29794-5-CD3-FaceQuality-231018}.  

Evaluated on MagFace~\cite{Meng-MagFace-CVPR-2021}, our findings reveal a decreasing trend of \textit{False Non-Match Rates (FNMR)} across all EDC curves when discarding facial expressions identified as furthest apart from neutrality. This observation highlights the significant impact of expression neutrality, specified as \textit{component quality element}~\cite{ISO-IEC-29794-5-CD3-FaceQuality-231018}, on the recognition performance. Further, the concatenation of intermediate feature layers with their Softmax values exhibits no measurable effect on the utility predictions. Similarly, the combination of \textit{HSE-1} and \textit{HSE-2} proves ineffective, except for a slight improvement with the AdaBoost classifier. 

Unexpectedly, SVMs outperform the remaining classifiers, as evidenced by the lowest pAUC values of $1.64\%$ achieved by HSE-2 and HSE-2-C. This observation is contrary to the findings on the classification task, underlining that a high classification accuracy of an algorithm cannot be projected to an equally good performance on predicting utility - a finding that aligns with~\cite{Grimmer-Neutrex-IJCB-2023}. A more detailed analysis of this phenomenon is presented in the next subsection, breaking down the class-specific discard flow and providing insights into how each classifier internally processes different expression types.

\subsection{Expression Class Analysis}
\label{sec:expression-class-analysis}

Finally, we explore the superiority of SVMs in comparison to the Random Forest and AdaBoost classifiers by discarding the lowest-utility facial images from the evaluation set while visualizing the expression class proportions. In this context, Figure~\ref{fig:stacked-area-plots} reveals that SVMs strategically reduce the proportions of \textit{surprised} and \textit{disgusted} expressions initially. The resulting positive effect on the recognition performance corresponds with the findings of~\cite{Grimmer-Neutrex-IJCB-2023}.    

In contrast, Random Forest and AdaBoost classifiers predominantly focus on diminishing the proportion of happy facial expressions. It is noteworthy that most FR systems are trained open-source datasets comprised of web-crawled facial images. Hence, they incorporate biases towards certain expression classes (\textit{e.g.}, smiling), neglecting underrepresented expressions like \textit{screaming} or \textit{yawning}. Consequently, classifying happy expressions as non-neutral and discarding them from the evaluation dataset leads to a subsequent decrease in recognition performance as the proportion of expression classes underrepresented in the training dataset increases. In conclusion, these findings underscore the need for system operators to customize the expression neutrality algorithm based on the application type while simultaneously considering potential biases inherent in the FR system.

%% file: sections/04-Conclusion.tex
\section{Conclusion}
\label{sec:conclusion}

In conclusion, our study addresses the issue of expression neutrality estimation in the context of component quality assessment for FR systems. We comprehensively benchmark several classifiers trained on features extracted from an efficient expression recognition model. Our findings indicate that SVMs are particularly well suited to predict utility, as they sort out facial images 1) deviating from expression neutrality and 2) causing a drop in recognition performance. In contrast, AdaBoost and Random Forest classifiers have been shown to significantly outperform SVM-based approaches, making them suitable algorithms for identifying non-neutral expressions. Overall, our work contributes efficient expression neutrality estimators, providing valuable insights for system optimization in biometric applications.

%% file: main.bbl
\begin{thebibliography}{10}

\bibitem{EU-Regulation-EES-InternalDocument-2017}
{Council of European Union}, ``{Council regulation ({EU}) no 2226/2017: Establishing an Entry/Exit System (EES)},'' 2017.

\bibitem{EU-Regulation-2017-2226-on-EES-171130}
{European Council}, ``Regulation 2017/2226 of the european parliament and of the council of 30 november 2017 on establishing an entry/exit system ({EES}) to register entry and exit data and refusal of entry data of third-country nationals,'' November 2017.

\bibitem{EU-ImplementingDecision-2019-329-on-EES-SampleQuality-190225}
{European Council}, ``Commission implementing decision 2019/329 of 25 february 2019 laying down the specifications for the quality, resolution and use of fingerprints and facial image for biometric verification and identification in the entry/exit system ({EES}),'' February 2019.

\bibitem{FRONTEX-BorderControl-BestPractices-InternalDocument-2015}
{Frontex}, ``{Best practice technical guidelines for Automated Border Control (ABC) systems},'' 2015.

\bibitem{schlett2022face}
T.~Schlett, C.~Rathgeb, O.~Henniger, J.~Galbally, J.~Fierrez, and C.~Busch, ``Face image quality assessment: A literature survey,'' {\em ACM Computing Surveys}, vol.~54, no.~10s, pp.~1--49, 2022.

\bibitem{EU-Regulation-2019-817-on-Interoperability-Framework-190520}
{European Council}, ``Regulation 2019/817 of the european parliament and of the council of 20 may 2019 on establishing a framework for interoperability between {EU} information systems in the field of borders and visa,'' May 2019.

\bibitem{ISO-IEC-29794-5-CD3-FaceQuality-231018}
{ISO/IEC JTC1 SC37 Biometrics}, {\em {ISO/IEC} {CD3} 29794-5 Information Technology - Biometric Sample Quality - Part 5: Face Image Data}.
\newblock International Organization for Standardization, 2023.

\bibitem{ISO-IEC-29794-1-QualityFramework-2023}
{ISO/IEC JTC1 SC37 Biometrics}, {\em {ISO/IEC} {DIS} 29794-1 Information Technology - Biometric Sample Quality - Part 1: Framework}.
\newblock International Organization for Standardization, 2023.

\bibitem{Savchenko-HSE-SISY-2021}
A.~V. Savchenko, ``Facial expression and attributes recognition based on multi-task learning of lightweight neural networks,'' in {\em Proc. of the 19th Intl. Symposium on Intelligent Systems and Informatics}, pp.~119--124, IEEE, 2021.

\bibitem{pena2021facial}
A.~Pe{\~n}a, A.~Morales, I.~Serna, J.~Fierrez, and A.~Lapedriza, ``Facial expressions as a vulnerability in face recognition,'' in {\em Intl. Conf. on Image Processing}, pp.~2988--2992, IEEE, 2021.

\bibitem{damer2018crazyfaces}
N.~Damer, Y.~Wainakh, V.~Boller, S.~von~den Berken, P.~Terh{\"o}rst, A.~Braun, and A.~Kuijper, ``Crazyfaces: Unassisted circumvention of watchlist face identification,'' in {\em 9th Intl. Conf. on Biometrics Theory, Applications and Systems}, pp.~1--9, IEEE, 2018.

\bibitem{ICAO-9303-p10-2021}
{International Civil Aviation Organization}, ``Machine readable passports -- part 10 -- logical data structure ({LDS}) for storage of biometrics and other data in the contactless integrated circuit ({IC}),'' 2021.
\newblock Last accessed: 2021-11-23.

\bibitem{ISO-IEC-39794-5-G3-FaceImage-191015}
{ISO/IEC JTC1 SC37 Biometrics}, {\em {ISO/IEC} 39794-5:2019 Information technology - Extensible biometric data interchange formats - Part 5: Face image data}.
\newblock International Organization for Standardization, 2019.

\bibitem{Grimmer-Neutrex-IJCB-2023}
M.~Grimmer, C.~Rathgeb, R.~Veldhuis, and C.~Busch, ``Neutrex: A 3d quality component measure on facial expression neutrality,'' {\em arXiv preprint arXiv:2308.09963}, 2023.

\bibitem{Tan-EfficientNet-PMLR-2021}
M.~Tan and Q.~Le, ``Efficientnet: Rethinking model scaling for convolutional neural networks,'' in {\em Intl. Conf, on Machine Learning}, pp.~6105--6114, PMLR, 2019.

\bibitem{Karras-CelebAHQ-arxiv-2017}
T.~Karras, T.~Aila, S.~Laine, and J.~Lehtinen, ``Progressive growing of gans for improved quality, stability, and variation,'' {\em arXiv preprint arXiv:1710.10196}, 2017.

\bibitem{Shi-CFD-TITS-2016}
Y.~Shi, L.~Cui, Z.~Qi, F.~Meng, and Z.~Chen, ``Automatic road crack detection using random structured forests,'' {\em IEEE Trans. on Intelligent Transportation Systems}, vol.~17, no.~12, pp.~3434--3445, 2016.

\bibitem{Lucey-CK+-CVPRW-2010}
P.~Lucey, J.~F. Cohn, T.~Kanade, J.~Saragih, Z.~Ambadar, and I.~Matthews, ``The extended cohn-kanade dataset (ck+): A complete dataset for action unit and emotion-specified expression,'' in {\em Conf. on Computer Vision and Pattern Recognition-Workshops}, pp.~94--101, IEEE, 2010.

\bibitem{GAN-FEAFA+-ICDIP-2022}
W.~Gan, J.~Xue, K.~Lu, Y.~Yan, P.~Gao, and J.~Lyu, ``{FEAFA+}: an extended well-annotated dataset for facial expression analysis and 3d facial animation,'' in {\em 14th Intl. Conf. on Digital Image Processing}, vol.~12342, pp.~307--316, SPIE, 2022.

\bibitem{Karras-FFHQ-CVPR-2019}
T.~Karras, S.~Laine, and T.~Aila, ``A style-based generator architecture for generative adversarial networks,'' in {\em Proc. of the IEEE/CVF Conf. on Computer Vision and Pattern Recognition}, pp.~4401--4410, 2019.

\bibitem{Phillips-FRGC-CVPR-2005}
P.~J. Phillips, P.~J. Flynn, T.~Scruggs, K.~W. Bowyer, J.~Chang, K.~Hoffman, J.~Marques, J.~Min, and W.~Worek, ``Overview of the face recognition grand challenge,'' in {\em Conf. on Computer Vision and Pattern Recognition}, vol.~1, pp.~947--954, IEEE, 2005.

\bibitem{Gross-MultiPie-IVC-2010}
R.~Gross, I.~Matthews, J.~Cohn, T.~Kanade, and S.~Baker, ``Multi-{PIE},'' {\em Image and Vision Computing}, vol.~28, no.~5, pp.~807--813, 2010.

\bibitem{Aifanti-MUG-WIAMIS-2010}
N.~Aifanti, C.~Papachristou, and A.~Delopoulos, ``The mug facial expression database,'' in {\em 11th Intl. Workshop on Image Analysis for Multimedia Interactive Services}, pp.~1--4, IEEE, 2010.

\bibitem{Meng-MagFace-CVPR-2021}
Q.~Meng, S.~Zhao, Z.~Huang, and F.~Zhou, ``{MagFace}: A universal representation for face recognition and quality assessment,'' in {\em Proc. of the IEEE/CVF Conf. on Computer Vision and Pattern Recognition}, pp.~14225--14234, 2021.

\end{thebibliography}
